\documentclass[runningheads]{llncs}
\usepackage{graphicx}
\usepackage{amsmath}
\usepackage{amssymb}
\usepackage{bm}
\usepackage{color}
\usepackage{multirow}
\usepackage{makecell}
\usepackage{soul}
\begin{document}
\title{How to Evaluate Your Dialogue Models:\\ A Review of Approaches}
\titlerunning{How to Evaluate Your Dialogue Models: A Review on Approaches}
%
\author{Xinmeng Li \and Wansen Wu \and Long Qin \and Quanjun Yin}
\authorrunning{Xinmeng Li, Wansen Wu, Long Qin and Quanjun Yin}
%
\institute{College of Systems Engineering, National University of Defense Technology, Changsha 410000, China \\
\email{xml.nudt@gmail.com}}
\maketitle              
\begin{abstract}
	Evaluating the quality of a dialogue system is an understudied problem. 
	The recent evolution of evaluation method motivated this survey, in which an explicit and comprehensive analysis of the existing methods is sought. 
	We are first to divide the evaluation methods into three classes, i.e., automatic evaluation, human-involved evaluation and user simulator based evaluation.
	Then, each class is covered with main features and the related evaluation metrics.
	The existence of benchmarks, suitable for the evaluation of dialogue techniques are also discussed in detail. Finally, some open issues are pointed out to bring the evaluation method into a new frontier.
\end{abstract}

\section{Introduction}
	Automatic evaluation is a non-trivial and challenging task in many sub-fields of natural language processing, e.g. text summarization~\cite{lin-2004-rouge}, machine translation~\cite{DBLP:conf/acl/PapineniRWZ02,DBLP:conf/acl/LinO04} and dialogue system~\cite{lowe2017towards}. Especially, the evaluation of dialogue system often poorly correlates with human judgement. This mismatch is a key bottleneck in migrating dialogue systems developed off-line for application in the real world~\cite{lowe2017towards}.	
	
	The goal of an evaluation method is to assess the performance of a system, which can be defined as ``the ability of a system to provide the function it has been designed for''~\cite{hastie2012metrics}.	
	Prior work often draws on the experience of evaluation methods in other natural language generation tasks, such as BLEU~\cite{DBLP:conf/acl/PapineniRWZ02}, NIST~\cite{DBLP:conf/acl/LinO04} and ROUGE-L~\cite{lin-2004-rouge}, which are widely used in machine translation, text summarization and image description. Take advantage of the progress of deep learning, efforts have done to learn evaluation metrics by neural networks~\cite{lowe2017towards,DBLP:conf/aaai/TaoMZY18,DBLP:conf/acl/MehriE20,DBLP:conf/acl/ZhaoLK20,DBLP:conf/iclr/ZhangKWWA20}. 

\begin{figure}[!htbp]
		\centering
		\vspace{-0.3cm}
		\includegraphics[width=1\linewidth]{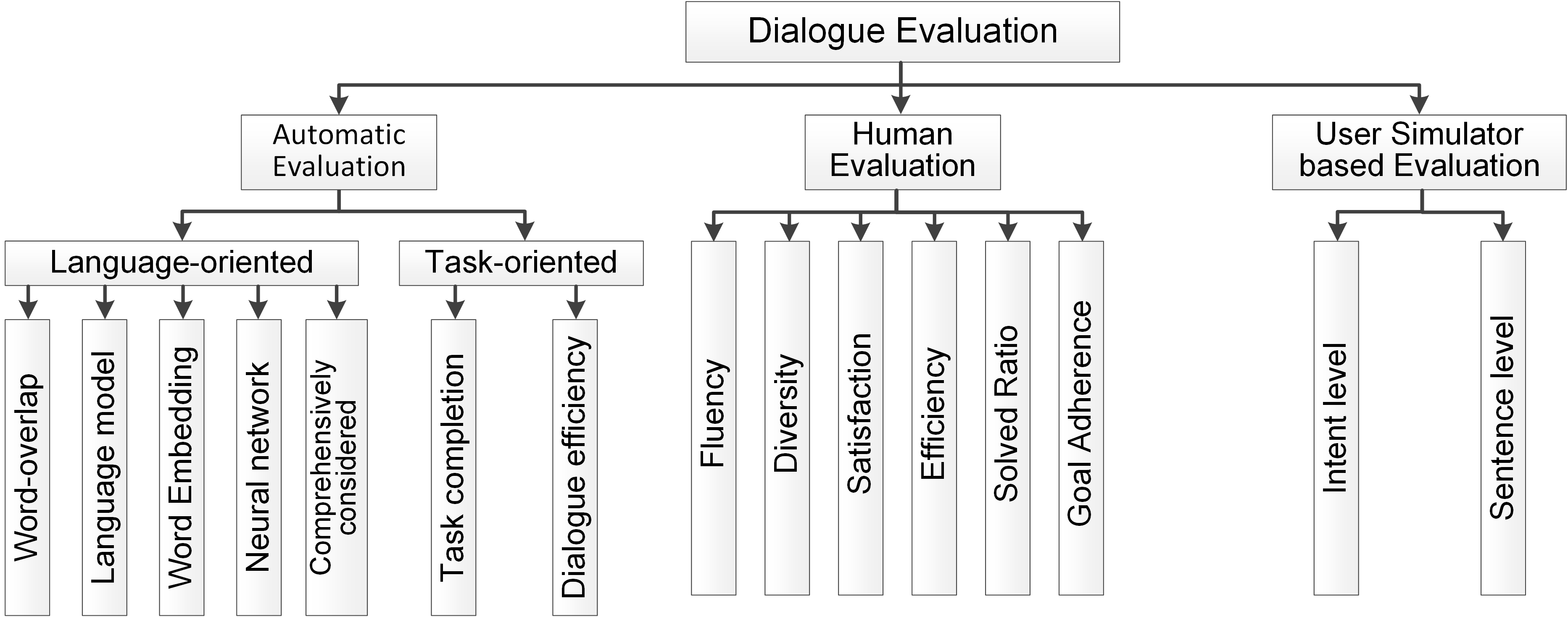}
		\vspace{-0.3cm}
		\caption{Taxonomy of the evaluation methods.}
		\label{fig:overview}
		\vspace{-0.3cm}
	\end{figure}

	
	In this article, we provide a thorough review of currently available dialogue evaluation paradigms. A method-based taxonomy of the existing
	evaluation metric is attempted here, including automatic corpus-based evaluation, human-involved evaluation and user simulator based evaluation. We give a condensed overview on Fig.\ref{fig:overview}, where an illustrative representation of the taxonomy is depicted. Deriu et al.\cite{DBLP:journals/corr/abs-1905-04071} has made a survey of evaluation methods for dialogue systems. In contrast to their work, we focus specifically on generative dialogue evaluation, and make a systematic review of existing techniques, including not only the state-of-the-arts but also those with latest trends. We further introduce how to take advantage of existing benchmarks.
	
\vspace{-0.3cm}
\section{Overview of Automatic Evaluation Methods}
\label{sec:automatic}
\vspace{-0.3cm}
	In this section, we introduce the automated approaches for evaluating dialogue systems by dividing them from two aspects, i.e., the language quality and task completion. The language-oriented evaluation reflects the informativity, coherence, fluency and grammaticality of the generated response. From another point of view, the task-oriented evaluation quantify the success rate and dialogue efficiency of a dialogue system.
\vspace{-0.3cm}
\subsection{Language-oriented evaluation}
	\subsubsection{Word-overlap based metric}
	If there is a standard response in corpus, the simple and natural way is to compare the word-overlap rate of the generated response with the gold one.
	BLEU~\cite{DBLP:conf/acl/PapineniRWZ02}, NIST~\cite{DBLP:conf/acl/LinO04} and METEOR~\cite{DBLP:conf/acl/BanerjeeL05} are initially developed for machine translation evaluation. BLEU computes geometric mean of the precision for n-gram, while NIST replaced geometric mean with arithmetic mean. METEOR considers precision and recall and identifies synonyms and paraphrases between the system output and the ground truth for more comprehensive matching. ROUGE-L~\cite{lin-2004-rouge} is a n-gram based recall which is first used in text summarization task. CIDEr~\cite{DBLP:conf/cvpr/VedantamZP15} synthetically considers TF-IDF weighting and n-grams averaging. All of them are frequently used word-overlap based metrics and are widely applied to dialogue system and natural language generation tasks~\cite{DBLP:conf/sigdial/NovikovaDR17,DBLP:conf/emnlp/WenGMSVY15}.
	Word-overlap based metrics correlate well with human judgements in lower diversity such as machine translation, but often fail in capturing the semantic similarity between the generated sentences and ground truths in dialogue systems for its diversity and dynamic nature. Finch et al.\cite{DBLP:conf/lrec/FinchAS04} argues that for most word-overlap metrics, a minimum of 4 references are needed in order to achieve reliable results. These metrics correlate better with human evaluation for datasets providing multiple ground truth sentences, which would lead to bias for reference sentence number~\cite{sharma2017relevance,DBLP:conf/sigdial/GuptaMZPEB19}. 

	Several variants of BLEU are also proposed to improve the evaluation correlation. Sun et al.\cite{DBLP:conf/acl/SunZ12} proposes iBLEU, a revised BLEU score to avoid trivial self-paraphrase and to measure the adequacy and diversity of the generated paraphrase sentence from SMT systems. 
	Galley et al.\cite{DBLP:conf/acl/GalleyBSJAQMGD15} proposes $\Delta$BLEU to evaluate conversational response generation task that admits a wide variety of possible outputs. However, upfront cost is needed to pay for human rating of the reference set. To remove the human intervention, $\upsilon$BLEU \cite{yuma2020ubleu} uses a neural network in place of human to annotates the reference set.
	\vspace{-0.3cm}
	\subsubsection{Language model based metric}
	Perplexity is widely used for measuring the performance of language models and also applied to conversational models~\cite{DBLP:journals/corr/VinyalsL15,shi2019build}. 
	Perplexity can give lower scores to similar dialogues, however the perplexity figure can be difficult to interpret as it ranges from 1 to infinity~\cite{DBLP:journals/ker/PietquinH13}. 
	The lexical diversity calculates the ratio of unique n-grams to total number of tokens in the dataset which expressing the variety of surface forms as opposed to repeating the same words and phrases~\cite{DBLP:conf/naacl/ShahHLT18}.		
	Distinct-1 and distinct-2 are applied to computed the number of distinct uni-grams and bi-grams divided by total number of generated words~\cite{DBLP:conf/naacl/LiGBGD16}. Vocabulary size and average utterance length can also be used to measure the language diversity~\cite{shi2019build}.
	\vspace{-0.3cm}
	\subsubsection{Embedding based metric}
	Embedding based metrics compute the similarity between of the predicted and the reference sentences by word embeddings. 
	Greedy Matching \cite{DBLP:conf/bea/RusL12} greedily matches each word in the candidate sentence to a word in the reference sentence based on the cosine similarity of their embeddings. Embedding average \cite{DBLP:journals/corr/WietingBGL15a} computes sentence-level embeddings by averaging the vector representations of their constituent words while Vector Extrema~\cite{forgues2014bootstrapping} takes the most extreme value of the vector representations of their constituent words for each dimension of the embedding. Different from aforementioned methods obtained through distributional word embeddings, skip-thought vector~\cite{DBLP:conf/nips/KirosZSZUTF15} is base on distributed sentence representations.  
	These metrics are appropriate to evaluate how semantically relevant and on-topic the responses are in evaluating the task of dialogue response generation~\cite{DBLP:conf/aaai/SerbanSLCPCB17}. However, embedding-based metrics only consist of basic averages of vectors obtained through distributional semantics, they are insufficiently complex for modeling sentence-level compositionality in dialogues~\cite{DBLP:conf/emnlp/LiuLSNCP16}. 
	To alleviate the problem, MoverScore~\cite{zhao2019moverscore} and SMS~\cite{clark2019sentence} further use sentence and word mover similarity for multi-sentence evaluating.
	\vspace{-0.3cm}
	\subsubsection{Neural network based metric}
	\label{sec:nn-based}
	Emerging trends are devoted to learning evaluation metrics by neural networks to get rid of the heuristic strategies \cite{lowe2017towards,DBLP:conf/aaai/TaoMZY18,DBLP:conf/iclr/ZhangKWWA20,sellam2020bleurt}. 
	Ryan et al.\cite{lowe2017towards} presents ADEM to predict human-like scores for dialogue responses. ADEM computes the score using a dot-product between the vector representations of the model response, the context and the reference response in a linearly transformed space to capture semantic similarity. 
	BERTScore~\cite{DBLP:conf/iclr/ZhangKWWA20} and BLEURT~\cite{sellam2020bleurt} take advantage of pre-trained language embeddings and computed the similarity score of two sentences as a sum of cosine similarities between their tokens’ embeddings.

	Reference-free evaluation methods are also suggested to get rid of ground truth sentences~\cite{DBLP:journals/corr/abs-1708-01759}. 
	An adversarial loss could be a way to directly evaluate the extent to which generated dialogue responses sound like they come from a human~\cite{DBLP:journals/corr/KannanV17}. 
	RUBER~\cite{DBLP:conf/aaai/TaoMZY18}, USR~\cite{DBLP:conf/acl/MehriE20} and RoBERTa-eval~\cite{DBLP:conf/acl/ZhaoLK20} considers both the referenced and unreferenced metrics with heuristic strategies to evaluate the response quality. 
	USR~\cite{DBLP:conf/acl/MehriE20}, an unsupervised and reference-free evaluation metric is proposed to address the shortcomings of standard metrics for language generation and shown strong correlation with human judgement.
	RoBERTa-eval \cite{DBLP:conf/acl/ZhaoLK20} investigates to use reference-free metrics, semi-supervised training, and pre-trained text encoders to reduce the bias with human judgement. 
	
	\vspace{-0.3cm}
	\subsubsection{Comprehensively considered metric}
	In addition to semantically coincidence, some other metrics such as fluency, topicality and grammaticality also are key factors and should be given attention in evaluation. 
	Readability and grammaticality are considered in \cite{DBLP:conf/emnlp/NovikovaDCR17} for sentence-level NLG evaluation. The evaluation is carried out on Flesch Reading Ease score~\cite{flesch1979how} and Stanford parser score\footnote{http://nlp.stanford.edu/software/parser-faq.shtml}. The results illustrate that word-based metrics show better correlations to human ratings of informativeness, whereas grammar-based metrics show better correlations to quality and naturalness. Dziri et al.\cite{DBLP:conf/naacl/DziriKMZ19} also includes entailment as an option to approximate dialogue coherence and quality.
	Guo et al.\cite{DBLP:journals/corr/abs-1801-03622} proposes to evaluate dialog quality with a series of topic-based metrics such as average topic depth, coarse topic breadth, topic keyword frequency and topic keyword coverage to evaluate the ability of conversational bots to lead and sustain engaging conversations on a topic and the diversity of topics the bot can handle. 
	The results show that a user's satisfaction correlated well with long and coherent on-topic conversations. 
\vspace{-0.3cm}
\subsection{Task-oriented evaluation}
\vspace{-0.3cm}
	Traditional task-oriented dialogue systems often adopt pipeline architecture~\cite{young2013pomdp}. There are metrics to evaluate performance of individual modules, such as intent accuracy for the natural language understanding, slot accuracy for dialogue state tracking, task success for policy learning and BLEU for natural language generation~\cite{takanobu2020empirical}. In addition, a comprehensive metric is needed to evaluate holistic system performance. 
	Generally, the task-oriented evaluation focuses on the task completion rate and dialogue efficiency of a dialogue system~\cite{DBLP:journals/nle/WalkerKL00}. 

	
	\vspace{-0.3cm}
	\subsubsection{Task completion}
	The task-oriented dialogue systems are developed to assist users in achieving specific goals. So the task completion rate, i.e., how well the dialogue system fulfills user's requirement, is of top priority to evaluate a dialogue system.
	Entity match rate evaluates task completion by determining if the system generates all correct constraints to search the indicated entities of the user~\cite{wen2017network}.
	Task success rate evaluates if the system answered all the associated information (e.g. phone number, ticket price)~\cite{wen2017network,lei-etal-2018-sequicity,shi2019build,madotto2018mem2seq}.
	Some fine distinctions of dialogue strategies can also be measured by inappropriate utterance ratio, turn correction ratio, concept accuracy, implicit recovery and transaction success~\cite{DBLP:conf/acl/HirschbergN96,danieli1995metrics}.
	
	\vspace{-0.3cm}
	\subsubsection{Dialogue efficiency}
	Dialogue efficiency measures the cost incurred in a dialogue, such as the dialogue length or the elapsed time. A direct yet effective metric is the turn number. 
			
	Reinforcement Learning based dialogue systems are mainly aimed to optimized task success rate and turn number \cite{DBLP:conf/ijcnlp/LiCLGC17,DBLP:conf/emnlp/LiMRJGG16,DBLP:conf/acl/WilliamsAZ17}, where the reward is shaped to better correlate with user satisfaction or be consistent with expert demonstrations directly~\cite{DBLP:conf/slt/LiHW14}. Success rate may only measure one aspect of the dialogue policy's quality. Focusing on information-seeking tasks, Ultes et al.\cite{DBLP:conf/interspeech/UltesBCMRSWGY17} proposes an interaction quality based reward to balance the dialogue policy. 
	
	Furthermore, success rate and turn number are also commonly used as system-level evaluation metrics for dialogue policy management and dialogue system performance~\cite{DBLP:conf/acl/SuGMRUVWY16,DBLP:conf/emnlp/PengLLGCLW17}. 

\vspace{-0.3cm}
\section{Human evaluation}
	\label{sec:human}
	\vspace{-0.3cm}
	Automated evaluation methods focus on single-turn quality, so they are often incapable to evaluate the holistic performance of dialogue system. Researchers have found that the actual system-level performances do not match the automatic evaluation results when interacting with real users~\cite{DBLP:journals/corr/abs-1801-06830}. 
	For practical use, a well-performed dialogue system is one which can interact efficiently and naturally with human subjects to complete an application-specific task~\cite{lowe2017towards}. So an ideal evaluation is to recruit human beings to interact with a dialogue agent to check whether it can successfully assist users to accomplish a given task. 
	Large-scale human evaluation is always performed in crowd-sourcing platform (i.e., CrowdFlower\footnote{faircrowd.work/platform/crowdflower}, Amazon Mechanical Turk\footnote{www.mturk.com}), where subjects interact with the dialogue systems and rate them via questionnaires \cite{DBLP:conf/sigdial/NovikovaDR17,DBLP:conf/aaai/SerbanSLCPCB17,DBLP:conf/naacl/ShahHLT18,DBLP:conf/acl/LiGBSGD16}.
	
	Human evaluation is a high subjective endeavor, so the measures often differentiate between research groups~\cite{DBLP:conf/inlg/HowcroftBCGHMMM20}.
	Anu et al.\cite{venkatesh2018evaluating} proposes a hybrid metric based on engagement, domain coverage, coherence, topical diversity and conversational depth. 
	See et al.\cite{DBLP:conf/naacl/SeeRKW19} designs two controllable neural text generation methods and conducted human evaluation to measure the effect of these control parameters from eight aspects: avoiding repetition, interestingness, making sense, fluency, listening, inquisitiveness, humanness and engagingness.	
	Ghandeharioun et al.\cite{DBLP:conf/nips/GhandehariounSJ19} uses a comprehensive evaluation strategy to approximate sentiment, semantic similarity, and engagement.
	Users are asked to give a rating on the naturalness, coherence, and task-completion capability of a system in \cite{li2018microsoft}.
	Shi et al.\cite{shi2019build} asks evaluators to interact with trained systems and obtained their opinions on satisfaction, efficiency, naturalness and rule-likeness and solved Ratio. Experimental results show that the auto-success is not necessarily correlated with the user-rated solved ratio. So human evaluation is indispensable to correct the automatic evaluation bias. 
	
	\begin{figure}[!htbp]
		\centering
		\vspace{-0.3cm}
		\includegraphics[width=1\linewidth]{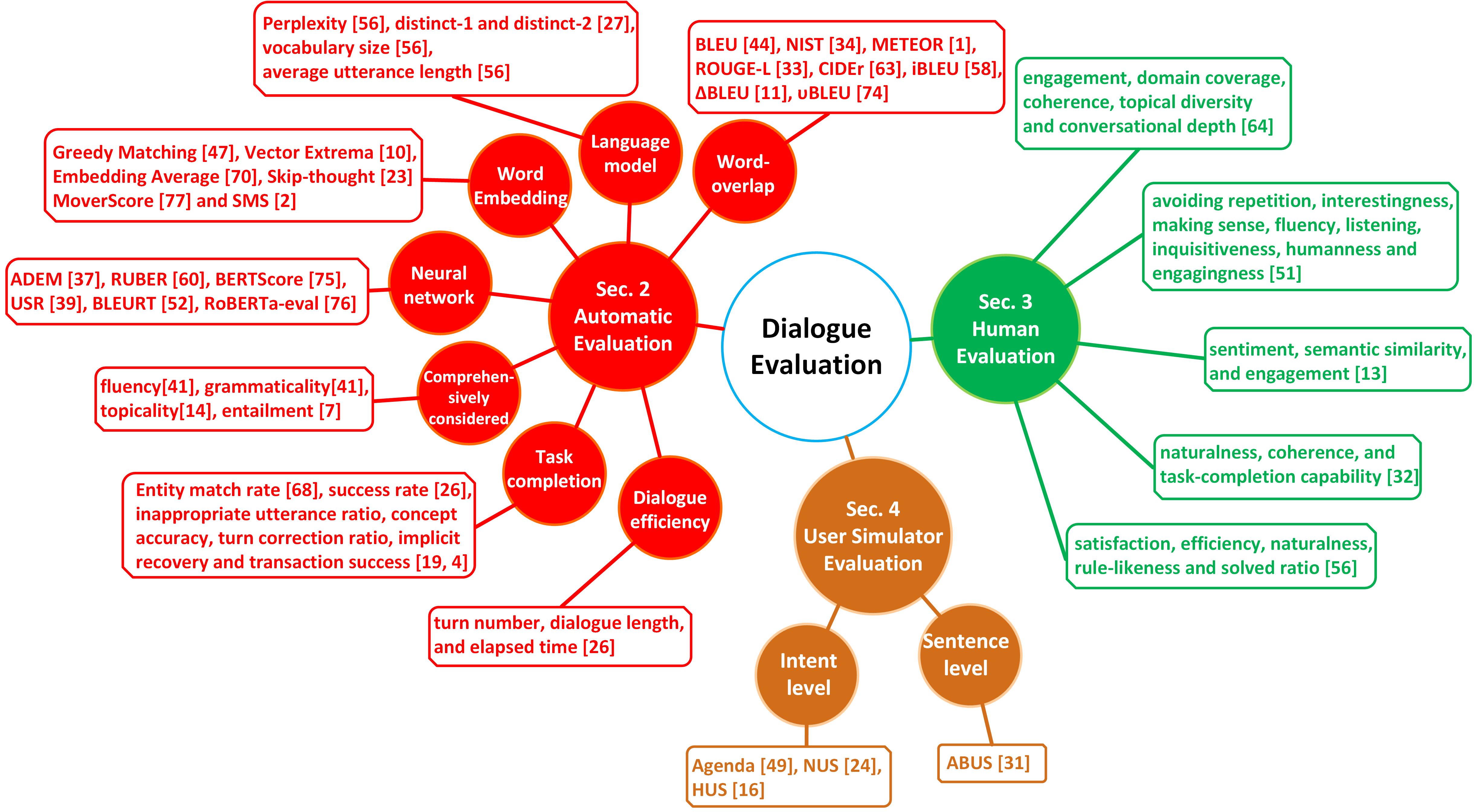}
		\vspace{-0.3cm}
		\caption{Summary of evaluation models and metrics. (See Section~\ref{sec:automatic},\ref{sec:human},\ref{sec:usersimulator} for Details)}
		\label{fig:overview-graph}
		\vspace{-0.3cm}
	\end{figure} 

	To balance the cost and accuracy, some hybrid evaluation methods are also proposed to leverage human and automatic evaluation, e.g, $\Delta$BLEU~\cite{DBLP:conf/acl/GalleyBSJAQMGD15}, HUSE~\cite{DBLP:conf/naacl/HashimotoZL19} and GENIE~\cite{khashabi2021genie}.

\vspace{-0.3cm}
\section{User simulator-based evaluation}
\label{sec:usersimulator}
\vspace{-0.3cm}
	Human evaluation is time-consuming and not scalable~\cite{sharma2017relevance}, while automatic evaluation metrics often lead to turn-level and system-level bias. To realize convenient and dynamic evaluation of dialogue systems, user simulators are needed to interact with the dialogue system.
	To this end, various efforts have been made to build user simulators which mimic human examiners to evaluate dialogue agents through interaction~\cite{schatzmann2006survey}.

	The commonly adopted method is to generate simulated dialogue by interacting with dialogue system and then assess the reality of the simulated dialogues.
	Accordingly, the evaluation can be categorized into intent-level and sentence-level. The intent-level focuses on dialogue policy part, interacting with dialogue systems through dialogue act or template-based utterances~\cite{schatzmann2007agenda,DBLP:conf/naacl/ShahHLT18,Gur2018UserMF}. Another line of work evaluates a complete task-oriented dialogue systems with fluent, dynamically generated natural language~\cite{DBLP:conf/ijcnlp/LiCLGC17}.
	L{\'{o}}pez{-}C{\'{o}}zar el al.\cite{DBLP:journals/air/Lopez-CozarCM06} tests the performance of spoken dialogue systems by artificially simulating the behavior of three types of user (very cooperative, cooperative and not very cooperative) to interact with previously developed dialogue systems. The user simulator enables the identification of problems relating to the speech recognition, spoken language understanding, and dialogue management components of the system. 	
	User simulators can also be used to perform cross validation to measure the miss-distance between automatic metrics with human evaluation~\cite{shi2019build}.	
	By employing a simulated user in a range of different experimental conditions, sufficient data can be generated to support a systematic analysis of potential problems and to enable fine-grained tuning of the system~\cite{DBLP:journals/air/Lopez-CozarCM06}.

	\begin{table}[!htpb]
	\small
	\centering
	\vspace{-0.3cm}
	\caption{\label{tab:benchmark} A summarization of the benchmarks. }
	\vspace{-0.3cm}
	\setlength{\tabcolsep}{1mm}{
		\begin{tabular}{|c|c|}
			\hline 
			Platform &  Feature  \\ \hline 
			PARADISE~\cite{walker-etal-1997-paradise}  &   \makecell[l]{comparing dialogue strategies by decoupling task requirements from \\ the dialogue behaviors} \\ \hline		
			PyDial~\cite{DBLP:conf/acl/UltesRSVKCBMWGY17} &  \makecell[l]{a dialogue policy specific platform} \\ \hline	
			ParlAI~\cite{miller2017parlai} &  \makecell[l]{end-to-end, multiple conversational tasks} \\ \hline			 
			\makecell[c]{ConvLab~\cite{DBLP:conf/acl/LeeZTZZLLPLHG19},\\ ConvLab-2~\cite{DBLP:conf/acl/ZhuZFLTLPGZH20}}  & \makecell[l]{supporting system-wise simulated evaluation and human evaluation} \\ \hline			
			Plato~\cite{papangelis2020plato} & \makecell[l]{single- or multi-party interactions, easy configuration and debug}  \\ \hline
			GENIE~\cite{khashabi2021genie} &  \makecell[l]{automate and standardize the human evaluation }  \\ \hline
			GEM~\cite{gehrmann2021the} & \makecell[l]{various tasks, multilingual, additional metrics allowed} \\ \hline			
	\end{tabular}}	
\end{table}

\vspace{-0.3cm}
\section{Benchmarks}
\vspace{-0.3cm}
	Given that evaluation method often performs differently across tasks and works, the benchmarks can act as testbed to evaluate the latest advances and facilitate the research frontier. We introduce the representative and prevalent benchmarks here, as listed in Table~\ref{tab:benchmark}.	

	PARADISE \cite{walker-etal-1997-paradise} is the first general evaluation framework for spoken dialogue systems. PARADISE supported comparisons among dialogue strategies by decoupling task requirements from the dialogue behaviors.
	PyDial\footnote{http://pydial.org} implements two success-based evaluator for dialogues evaluation, including a objective success evaluator to compare the constraints and requests the system identifies with the true values, and a subjective task success evaluator to queries the user about the outcome of the dialogue~\cite{DBLP:conf/acl/UltesRSVKCBMWGY17}.
	ParlAI\footnote{http://parl.ai} provides a unified framework with various conversational tasks, including dialogue system, reading comprehension, etc~\cite{miller2017parlai}.
	Convlab\footnote{http://convlab.github.io} provides a platform for researchers to develop dialogue systems with various architectures or configurations. It supports system-wise simulated evaluation and also provides the interface of AMT platform for human evaluation~\cite{DBLP:conf/acl/LeeZTZZLLPLHG19}. ConvLab-2\footnote{https://github.com/thu-coai/ConvLab-2} is the continuator of ConvLab with more powerful architectures and supporting more datasets~\cite{DBLP:conf/acl/ZhuZFLTLPGZH20}.
	On the other hand, Plato\footnote{https://github.com/uber-research/plato-research-dialogue-system} is a abstraction level platform, which is well-designed for easy of understanding and debugging for conversational agents~\cite{papangelis2020plato}. 
	GENIE\footnote{https://genie.apps.allenai.org} provides a leaderboard for standardizing the human evaluation on text generation systems~\cite{khashabi2021genie}.  
	GEM\footnote{https://gem-benchmark.com/} provides a benchmark integrated with a wide set of tasks and allows the integration of additional metrics to identify the gaps and then prioritize the direction for improvement~\cite{gehrmann2021the}. 
\vspace{-0.3cm}
\section{Open issues and questions}
\vspace{-0.3cm}
	Evaluation is a non-trivial and understudied task. Besides common challenges inherent in the existing evaluation methods, there are some special challenges on evaluation of the dialogue system. 

\vspace{-0.3cm}
\subsection{Reference-free evaluation}
	Generally, a dialogue system is trained to imitate the ground truth response from dataset. Most evaluation methods compare the generated response with the ground truth. However, there are a diverse of potential responses to a question in a common conversation. 
	\cite{sharma2017relevance} points out that multiple ground truth sentences are helpful to improve the correlation between metrics and human evaluation for which provide more adequate and diversified semantic information. However, datasets collecting is also a time-consuming and laborious work. 
	As stated in Section~\ref{sec:nn-based}, the reference-free evaluation method has become a new trend to improve the scalability and generalization ability of evaluation~\cite{DBLP:conf/aaai/SaiGKS19,DBLP:conf/acl/ZhaoLK20}.
\vspace{-0.3cm}	
\subsection{Interactively dynamic evaluation}
	Generally, a conversation is a multi-turn interactive process between the user and the dialogue system. In prior evaluation method, the gold dialogue history (standard user utterance and dialogue response from corpus) is fed to dialogue system in each turn without considering the actually generated response before~\cite{wen2017network,madotto2018mem2seq,wu2019global}.  Without dynamic interaction, the errors occurred in prior turn would be buried and can't be passed to next turn. That is to say, the turn-level evaluation can't reflect the system-level performance of dialogue system.
	So dynamic evaluation method is more effective to strengthen the error checking and recovery mechanisms for dialogue system. 
	
\vspace{-0.3cm}	
\subsection{User simulator with fluent natural language}
	Existing user simulators only focus on dialogue policy part, interacting with dialogue systems through well-structured formal languages~\cite{shi2019build}. Such user simulators are insufficient to evaluate a complete task-oriented dialogue systems, as they cannot appropriately assess fluent, dynamically generated natural language.
	With the recent wave of end-to-end dialogue systems ~\cite{wen2017network,lei-etal-2018-sequicity,madotto2018mem2seq,wu2019global}, the community starts to emphasize on natural utterances and realistic usage of dialogue systems. It would push researchers to build an effective user simulator which realistically evaluate dialogue agents through interacting with them using fluent natural languages.
\vspace{-0.3cm}
\section{Conclusion}
	In this paper, we summarize the evaluation methods of dialogue systems and a method-based taxonomy of the existing evaluation metric is attempted here.
	Even though much work has done to construct an effective evaluation method, it is still a challenging and understudied work to capture all aspects of dialogue response from naturalness and coherence to long-term engagement and flow. 
	Finally, we believe that the redefinition of user simulator is the most promising direction to build an ideal evaluator.

\bibliography{anthology}
\bibliographystyle{splncs04}

%
%
%
%
\end{document}